\documentclass[letterpaper]{article} %
\usepackage[]{aaai23}  %
\usepackage{times}  %
\usepackage{helvet}  %
\usepackage{courier}  %
\usepackage[hyphens]{url}  %
\usepackage{graphicx} %
\usepackage{natbib}  %
\usepackage{caption} %
\usepackage{algorithm}

\usepackage{newfloat}
\usepackage{listings}
\DeclareCaptionStyle{ruled}{labelfont=normalfont,labelsep=colon,strut=off} %
\floatstyle{ruled}
\newfloat{listing}{tb}{lst}{}
\floatname{listing}{Listing}
\title{Plausibility-Based Heuristics for Latent Space Classical Planning}

\author{
  Yuta Takata \textsuperscript{\rm 1}
  Alex Fukunaga \textsuperscript{\rm 2}\\
}
\affiliations{
  \textsuperscript{\rm 1} Fabalis
  \textsuperscript{\rm 2} The University of Tokyo

}

\usepackage{comment}
\usepackage{xspace}
\usepackage{bm}

\newcommand{\Xtr}{\mathcal{X}}

\usepackage{subcaption}
\usepackage{multirow}
\usepackage{algorithm}
\usepackage{amsmath}
\usepackage{amsfonts}
\usepackage{amsthm}
\usepackage[noend]{algpseudocode}
\usepackage{booktabs} %
\usepackage{xr} 
\usepackage{setspace}
\usepackage{amssymb} %
\theoremstyle{definition}
\newtheorem{def.}{Definition}
\newtheorem{th.}{Theorem}
\newtheorem{lem.}{Lemma}
\newtheorem{cor.}{Corollary}
\newtheorem{obs.}{Observation}

\newif\ifwastar
\wastarfalse
\newcommand{\astar}{$\mathit{A^*}$\xspace}

\newcommand{\latplan}{LatPlan\xspace}
\newcommand{\mands}{M\&S}
\newcommand{\chisquare}{$\mathit{\chi^2}$}

\newcommand{\hchisquare}{$h_{\mathit{\chi^2}}$}
\newcommand{\hkldivergence}{$h_{\mathit{KL}}$\xspace}

\newcommand{\coptimal}{$c$-optimal\xspace}

\nocopyright

\begin{document}

\maketitle

\begin{abstract} 
Recent work on \latplan has shown that it is possible to learn models for
domain-independent classical planners from unlabeled image data.
Although PDDL models acquired by \latplan can be solved using standard PDDL planners, the resulting latent-space
plan may be invalid with respect to the underlying, ground-truth domain (e.g., the latent-space plan may include hallucinatory/invalid states).
We propose Plausibility-Based Heuristics, which are domain-independent plausibility metrics which can be computed for each state evaluated during search and uses 
as %
a heuristic function for best-first search.
We show that PBH significantly increases the number of valid found plans on
image-based tile puzzle and Towers of Hanoi domains.

\end{abstract}

\section{Introduction}
Automated acquisition of domain models for planners from subsymbolic data (e.g., images taken by cameras) poses an important challenge for autonomous systems that must operate in new, unknown environments.
Recent work has shown that it is possible to learn domain models for
domain-independent classical planners from unlabeled, unannotated pairs of images
representing the state of the world before and after actions are executed \cite{DBLP:conf/aaai/AsaiF18,AsaiKFM22}.
\latplan uses deep learning to learn a {\it latent} propositional representation of the domain, which can be output as a STRIPS PDDL model.
These latent propositions as well as the action model are learned from scratch, without domain-specific prior knowledge, i.e., \latplan grounds the symbols \cite{Steels2008} and learns an action model with respect to the latent propositions.
Problem instances for these learned models can be solved using standard search-based planners using standard domain-independent heuristics. 

\begin{figure}[tb]
  \centering
  \includegraphics[width=1.0\linewidth]{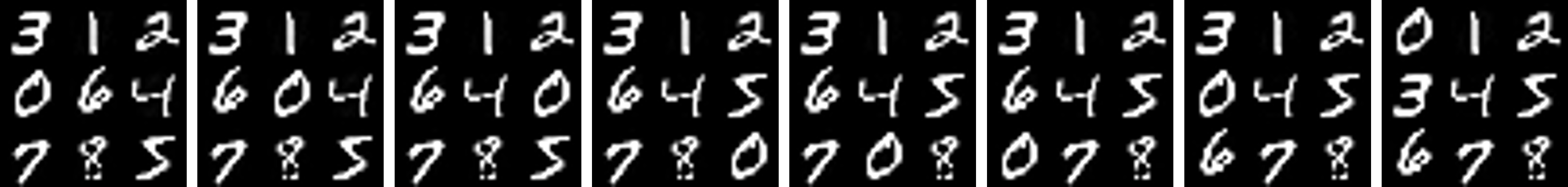}
  \includegraphics[width=1.0\linewidth]{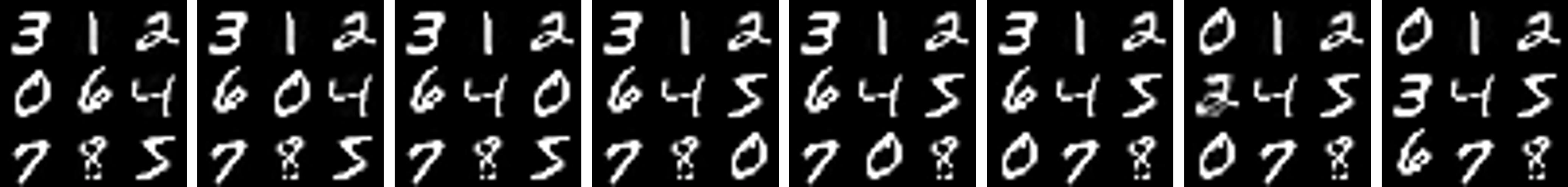}
 \caption{ 
 Visualizations of 2 latent plans for MNIST-8 puzzle (start: left $\rightarrow$ goal:right).
 The top plan is valid and optimal. %
 The bottom plan is invalid  because the 7th image has two ``0''s and an unclear digit which could be interpreted as a 2, 3, or 6.
 }
 \label{fig:mnist8-valid-and-invalid}
\end{figure}

However, an inherent issue with learned models is that propositions and actions in the learned, latent PDDL model do not necessarily correspond to those of the underlying problem.
Thus, a significant problem is that although standard search algorithms such as Fast Downward can find solutions which are correct with respect to the latent PDDL model, these solutions may not be valid with respect to the underlying problem.
For example, Figure \ref{fig:mnist8-valid-and-invalid} shows an invalid plan found by \latplan for the image-based MNIST-8 puzzle.

Previous work focused on improving the architecture and learning objectives so that models learned by \latplan accurately correspond to the underlying domains
\cite{DBLP:conf/aips/AsaiK19,AsaiKFM22}.
Despite these efforts,
the current implementation of \latplan has limitations.
Although \latplan can successfully solve image-based 15-puzzles,
it has not been able to find valid solutions for a 4-disk Towers of Hanoi problem, even though it is much ``simpler'' than the 15-puzzle  with respect to the size of the search space of the underlying problem \citep[Appendix M]{AsaiKFM22}.

In this paper, we investigate a complementary approach.
Instead of the learning component, we focus on the search component of \latplan.
Previous work relied on using unmodified, standard planner (Fast Downward) to solve the PDDL instances generated by \latplan.
We improve the ability to find valid plans by making \latplan-specific (but domain-independent) extensions to the search component.

We propose Plausibility-Based Heuristics (PBH), which
uses the learned \latplan model during the search to evaluate the {\it plausibility} of a state, i.e., the likelihood that the state is valid with respect to the underlying domain.
The plausibility measure can be used as the
heuristic evaluation function $h$ 
for best-first search (e.g., \astar, GBFS).
Instead of guiding search according to standard distance-based heuristics,
we prioritize search according to plausibility, as finding a valid plan takes precedence over finding short but invalid plans.
This is somewhat similar to the idea of using novelty and width-based exploration in black-box domains where standard heuristics do not lead to a solution \cite{DBLP:conf/ijcai/LipovetzkyRG15}.

As \latplan is a domain-independent, image-based domain learner/planner,
an important goal of this work is to identify useful, {\it domain-independent} plausibility metrics which are useful on some subset of \latplan domains.
We propose image-based plausibility measures which can be computed {\it during search} from the propositional latent state.

The remainder of the paper is structured as follows. Section \ref{sec:preliminaries} reviews \latplan.
Section \ref{sec:validity} discusses the validity of latent space plans,
and show that the ability to find valid plans is highly sensitive to \latplan hyperparameter settings.
  Section \ref{sec:plausibility-metrics} proposes metrics for the plausibility of states in latent search spaces, and their use as heuristic functions for best-first search.
  Section \ref{sec:experiments} experimentally evaluates PBH and shows 
that
it significantly improves the number of valid solutions found by the search algorithm.
Section \ref{sec:conclusion} concludes with a discussion of our results.

\section{Preliminaries and Background}
\label{sec:preliminaries}

A STRIPS planning task \cite{DBLP:journals/ai/FikesN71} is defined by a 4-tuple $T=\langle P, A, I, G \rangle$. 
$P$ is a set of propositions.
$A$ is a set of actions.
$I \subset P$ represents the initial state, and $G \subset P$ is the goal condition.
A state is represented by a subset of $P$, and applying an action to a state adds some propositions and removes some propositions in the state.
Each action $a \in A$ is composed of three subsets of $P$,$\langle \mathit{pre(a)},\mathit{add(a)},\mathit{del(a)} \rangle$, which are the
{\it preconditions}, {\it add effects}, and {\it delete effects}.
An action $a$ is applicable to a state $S$ iff $pre(a) \subseteq S$.
By applying $a$ to $S$, propositions in $S$ change to $S(a) = ((S \setminus \mathit{del(a)}) \cup \mathit{add(a)})$.
For a sequence of actions $\pi= (a_0,\ldots, a_n)$, we use $S(\pi)$ to denote
$((((S \setminus \mathit{del(a_0)}) \cup \mathit{add(a_0)}) \setminus \mathit{del(a_1)}) \cup \cdots) \cup \mathit{add(a_n)}$.
A solution to $T$ is a sequence of actions that transforms $I$ to a state $S$ satisfying $G \subseteq S$.
A feasible solution, i.e., a plan, is a sequence of actions
$ \pi= (a_0,\ldots, a_n)$ that satisfies
(i) $\forall i,\mathit{pre(a_i)} \subseteq I((a_0, \ldots ,a_{i-1}))$ 
and (ii) $G \subseteq I(\pi)$.

\subsubsection{Latent-space planning using \latplan}

Given
only an unlabeled set of image pairs showing a subset of transitions allowed in the environment
(training inputs), \latplan learns a complete propositional PDDL action model of the environment.
Later, when a pair of images representing the initial and the goal states (planning inputs) is given,
\latplan finds a plan to the goal state in a symbolic latent space and returns a visualized plan execution.

Assume that the environment poses a STRIPS planning problem.
Assume that images (e.g., images taken using a camera) are represented as a vector of integers.
Let $\cal{O}$ be the space of observations, where
an {\it observation} $o \in \cal{O}$ is an image representing a state of the environment.

\begin{figure}[bt]
 \centering
 \includegraphics[width=0.65\linewidth]{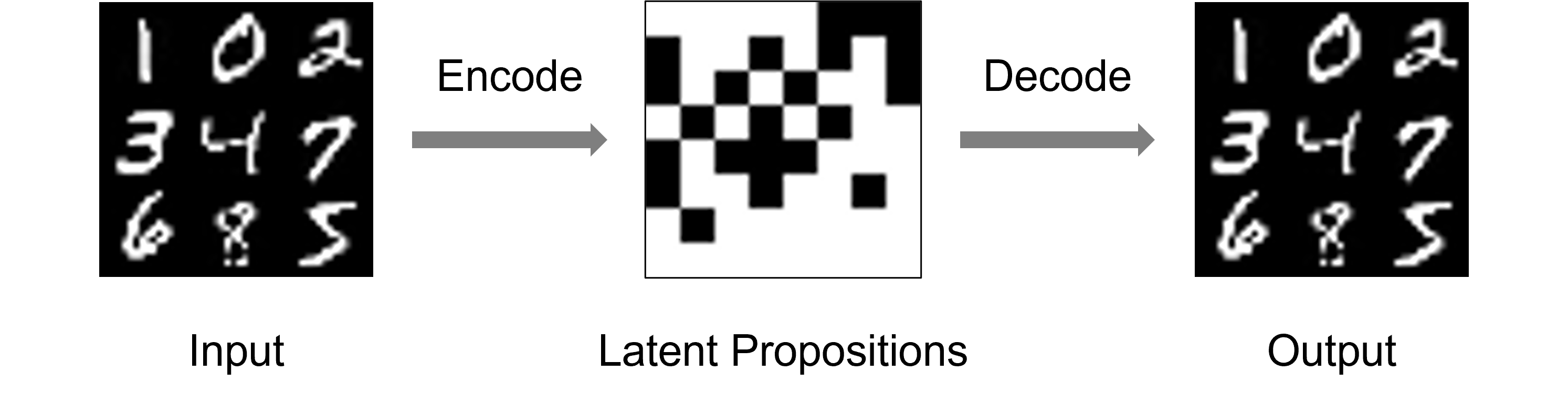}
 \caption{ %
   State Autoencoder
 }
 \label{fig:autoencoder}
\end{figure}

The input to \latplan is 
$\Xtr$, a set of pairs of images. %
Each image pair $(\bm{x}_{i,\textit{before}}, \bm{x}_{i,\textit{after}}) \in \Xtr$
represents a transition from an observation  $\bm{x}_{i,\textit{before}}$ to another observation  $\bm{x}_{i,\textit{after}}$ caused by an unknown action.
Based only on $\Xtr$,
\latplan learns two functions: 
(1) $\mathit{Encode}(o) : \cal{O} \to$ $(0,1)^{N}$,
which maps an image $o$ to a propositional vector,
and
(2) $\mathit{Decode}(o) : (0,1)^N \to \cal{O}$,
which maps a propositional vector to an image (Fig. \ref{fig:autoencoder}).
$N$ is a hyperparmeter for the number of latent propositions.
\latplan learns a Discrete Variational Autoencoder (DVAE) which perform both $\mathit{Encode}$ and $\mathit{Decode}$. Unlike a standard autoencoder \cite{hinton2006reducing}
where unit activations are continuous, the DVAE uses a binary (propositional) activation \cite{DBLP:conf/iclr/JangGP17,DBLP:conf/iclr/MaddisonMT17}.
The propositional output/input of  $\mathit{Encode}$/$\mathit{Decode}$ is called the {\it latent representation}.

Using $\mathit{Encode}$ and $\mathit{Decode}$, and $\Xtr$, 
\latplan generates a {\it latent domain} $D^l=\langle P^l, A^l \rangle$,
where $P^l$ is a set of (latent) propositions and
$A^l$ is a set of actions in latent space. 
Each action $a^l \in A^l$ is composed of three subsets of $P^l$,
$\langle \mathit{pre(a^l)},\mathit{add(a^l)},\mathit{del(a^l)} \rangle$,
and action $a^l$ is applicable to a state $S^l$ iff $pre(a^l) \subseteq S^l$.
By applying $a^l$ to $S^l$, the propositions in $S^l$ change to $S^l(a^l) = ((S^l \setminus \mathit{del(a^l)}) \cup \mathit{add(a^l)})$.

Let $o_I$ be an image of a start state in the environment (e.g., jumbled 8-puzzle configuration).
Let $o_G$ be an image representing a desired state of the environment (e.g., solved 8-puzzle state).
A latent STRIPS planning task
 $T^l=\langle P^l, A^l, I^l, G^l \rangle$, where
$I^l = \mathit{Encode(o_I)}$ and $G^l = \mathit{Encode(o_G)}$.
A solution $ \pi^l= (a^l_0,\ldots, a^l_n)$ to $T^l$ is a sequence of actions that transforms $I^l$ to a state $S^l$ that satisfies
$G^l \subseteq P^l$.
As $T^l$ is just a standard STRIPS task, $T^l$ can be solved using
any planner which takes PDDL input.

The latent space plan $ \pi^l$ is a sequence of actions in the latent representation, and can not be directly interpreted and executed by the agent.
Therefore, \latplan uses its learned action model and $\mathit{Decode}$ to map the latent states visited by $\pi^l$ to a sequence of images $(\bm{y}_0,\ldots,\bm{y_n})$, called the {\it visualization} of $\pi^l$, 
 which can be interpreted and executed by the agent, e.g., a human or robot can move the sliding tile puzzle tiles according to the sequence of images.

 \section{Validity of Latent Space Plans}
 \label{sec:validity}
An inherent issue with domain learning systems is that the learned domain may differ from the underlying, canonical domain.  
The latent domain $D^l$ learned by \latplan can contain many states/actions which do not correspond to any valid canonical state/actions.
For example, although the MNIST-8 puzzle corresponds to a canonical 8-puzzle with a branching factor $b$ of at most 4,
the search space of the learned MNIST-8 latent space domain used in the Section \ref{sec:experiments} experiments has a maximum (over all expanded nodes) $b$ of 124.9 (mean of 20 instances).
Thus, the search algorithm may incorrectly find latent search space plans which can not be executed.

We define the following notion of validity of a plan generated by \latplan,
relative to some target STRIPS task $T$ which the latent domain learned by \latplan is intended to correspond to.

\begin{def.}
  We say that $\pi^l$ is {\it valid} with respect to a canonical STRIPS task $T=\langle P, A, I, G \rangle$ if:
  for each image $\bm{y}_i$ in the visualization of $\pi^l$ except the goal,
  $\bm{y}_i$ corresponds to a state $S_i \subset P$,
  $\bm{y}_{i+1}$ corresponds to a valid state $S_{i+1} \subset P$,
  and $\exists a \in A$ such that $S_i(a) = S_{i+1}$.
\end{def.}

Similarly, length optimality of $\pi^l$ wrto $T^l$ does not necessarily imply optimality wrto the  underlying canonical task $T$, so we define the following condition which is satisfied when $\pi^l$ is optimal wrto $T$.

\begin{def.} We say that $\pi^l$ is {\it $c$-optimal} relative to $T$ if 
  $\pi^l$ is valid and  $|\pi^l|$ is equal to that of the minimal-length plan for $T$.
\end{def.}

In Figure \ref{fig:mnist8-valid-and-invalid}, the top plan is valid and \coptimal. The bottom plan is invalid, because the 7th image from the left has two ``0''s and an unclear digit.

In the experiments below,
the \latplan input $\Xtr$ is generated by a 
domain-specific generator which has full knowledge of a canonical STRIPS domain (e.g., 8-puzzle), and
outputs a set of valid sample image pairs $\Xtr$ for that domain. The start/goal images $o_I$ and $o_G$ are also generated using this generator.
Therefore, for each \latplan domain, plan validity for a latent space plan $\pi^l$ 
is checked using a domain-specific validator.

Since our experiments use validation code to test latent plan validity,
one might wonder whether we could simply use this validator during search to prune invalid states. %
However, unlike domain-independent validators for PDDL domains such as VAL \cite{DBLP:conf/ictai/HoweyLF04}, the \latplan image-based plan validators are carefully tuned, hand-coded domain-specific tools (see \citet[Appendix L.1]{AsaiKFM22}),
which use extensive domain knowledge
e.g., for MNIST-8, the knowledge that a valid state image is composed of 8 unique tiles, each of which corresponds to a portion of the unscrambled image.
If we have such knowledge, then instead of a weak learning model such as LatPlan, a model learning method which can exploit such prior knowledge would probably be a better alternative. 
Therefore, such domain-dependent image validators are unlikely to be available in domains for which \latplan is a viable approach.

\subsection{Plan Validity and \latplan Model Learning Hyperparameters }
\begin{table*}[tb]
\centering
\resizebox{\textwidth}{!}{%
\begin{tabular}{c|c|ccccccccc|ccccccccc}
 &
  $\epsilon$ &
  \multicolumn{9}{c|}{0.1} &
  \multicolumn{9}{c}{0.5} \\ \hline
 &
  $N$ &
  \multicolumn{3}{c|}{50} &
  \multicolumn{3}{c|}{100} &
  \multicolumn{3}{c|}{300} &
  \multicolumn{3}{c|}{50} &
  \multicolumn{3}{c|}{100} &
  \multicolumn{3}{c}{300} \\ \hline
$\beta_3$ &
  $\beta_1$ &
  found &
  valid &
  \multicolumn{1}{c|}{$c$-optimal} &
  found &
  valid &
  \multicolumn{1}{c|}{$c$-optimal} &
  found &
  valid &
  $c$-optimal &
  found &
  valid &
  \multicolumn{1}{c|}{$c$-optimal} &
  found &
  valid &
  \multicolumn{1}{c|}{$c$-optimal} &
  found &
  valid &
  $c$-optimal \\ \hline
\multirow{2}{*}{1} &
  1 &
  20 &
  9 &
  \multicolumn{1}{c|}{9} &
  20 &
  19 &
  \multicolumn{1}{c|}{19} &
  20 &
  20 &
  19 &
  20 &
  13 &
  \multicolumn{1}{c|}{13} &
  20 &
  19 &
  \multicolumn{1}{c|}{19} &
  20 &
  20 &
  20 \\
 &
  10 &
  20 &
  4 &
  \multicolumn{1}{c|}{4} &
  3 &
  3 &
  \multicolumn{1}{c|}{3} &
  20 &
  20 &
  20 &
  20 &
  7 &
  \multicolumn{1}{c|}{7} &
  2 &
  0 &
  \multicolumn{1}{c|}{0} &
  0 &
  0 &
  0 \\ \hline
\multirow{2}{*}{10} &
  1 &
  20 &
  8 &
  \multicolumn{1}{c|}{8} &
  17 &
  17 &
  \multicolumn{1}{c|}{17} &
  20 &
  20 &
  19 &
  20 &
  4 &
  \multicolumn{1}{c|}{4} &
  18 &
  16 &
  \multicolumn{1}{c|}{16} &
  18 &
  18 &
  13 \\
 &
  10 &
  20 &
  0 &
  \multicolumn{1}{c|}{0} &
  20 &
  11 &
  \multicolumn{1}{c|}{11} &
  20 &
  20 &
  20 &
  20 &
  6 &
  \multicolumn{1}{c|}{6} &
  1 &
  0 &
  \multicolumn{1}{c|}{0} &
  0 &
  0 &
  0 \\ \hline
\multirow{2}{*}{100} &
  1 &
  20 &
  15 &
  \multicolumn{1}{c|}{15} &
  20 &
  9 &
  \multicolumn{1}{c|}{8} &
  11 &
  11 &
  0 &
  20 &
  9 &
  \multicolumn{1}{c|}{9} &
  20 &
  19 &
  \multicolumn{1}{c|}{19} &
  0 &
  0 &
  0 \\
 &
  10 &
  20 &
  4 &
  \multicolumn{1}{c|}{4} &
  20 &
  5 &
  \multicolumn{1}{c|}{5} &
  0 &
  0 &
  0 &
  20 &
  10 &
  \multicolumn{1}{c|}{10} &
  0 &
  0 &
  \multicolumn{1}{c|}{0} &
  0 &
  0 &
  0 \\ \hline
\multirow{2}{*}{1000} &
  1 &
  20 &
  2 &
  \multicolumn{1}{c|}{2} &
  0 &
  0 &
  \multicolumn{1}{c|}{0} &
  0 &
  0 &
  0 &
  20 &
  11 &
  \multicolumn{1}{c|}{11} &
  0 &
  0 &
  \multicolumn{1}{c|}{0} &
  0 &
  0 &
  0 \\
 &
  10 &
  20 &
  4 &
  \multicolumn{1}{c|}{4} &
  0 &
  0 &
  \multicolumn{1}{c|}{0} &
  5 &
  5 &
  0 &
  20 &
  11 &
  \multicolumn{1}{c|}{11} &
  0 &
  0 &
  \multicolumn{1}{c|}{0} &
  0 &
  0 &
  0 \\ \hline
\multirow{2}{*}{10000} &
  1 &
  12 &
  0 &
  \multicolumn{1}{c|}{0} &
  6 &
  0 &
  \multicolumn{1}{c|}{0} &
  0 &
  0 &
  0 &
  5 &
  0 &
  \multicolumn{1}{c|}{0} &
  7 &
  2 &
  \multicolumn{1}{c|}{2} &
  0 &
  0 &
  0 \\
 &
  10 &
  12 &
  1 &
  \multicolumn{1}{c|}{1} &
  4 &
  0 &
  \multicolumn{1}{c|}{0} &
  0 &
  0 &
  0 &
  12 &
  0 &
  \multicolumn{1}{c|}{0} &
  7 &
  0 &
  \multicolumn{1}{c|}{0} &
  0 &
  0 &
  0
\end{tabular}%
}
\caption{
  Hyperparameter sweep study of \latplan (AMA4+ action model) on MNIST 8-puzzle.
  A latent domain was learned for each all combinations of $(N, \epsilon, \beta_1, \beta_3)$ for $ N \in \{50, 100, 300\},\ \epsilon \in \{0.1, 0.5\},\ \beta_1 \in \{1, 10\},\  \beta_3 \in \{1, 10, 100, 1000, 10000\}$), where 
$N = |P^l|$, the number of latent propositional variables , 
$\epsilon$ is the parameter for the Bernoulli($\epsilon$) distribution used as the prior for the latent random variables in the Binary-Concreate VAE, and
  $\beta_1$ and $\beta_3$ are the regularization parameters for the KL-divergence used in the ELBO loss function. %
  For each domain, 20 instances were solved using Fast Downward (\astar with blind heuristic).
  }
\label{tab:hyperparameters}
\end{table*}

\label{sec:hyperparameter-sensitivity}

\latplan has 4 key hyperparameters related to model learning: 
(1) The number of latent propositional variables $N = |P^l|$
(2) $\epsilon$, the parameter for the Bernoulli($\epsilon$) distribution used as the prior for the latent random variables in the Binary-Concrete VAE \cite[Sec 5.4]{AsaiKFM22}.
(3,4) the regularization parameters $\beta_1$ and $\beta_3$ for the KL-divergence used in the Evidence Lower Bound (ELBO) loss function. %
The best values for these hyperparameters are domain-dependent.
Asai et al (\citeyear{AsaiKFM22}) showed that the accuracy of the learned models depended significantly on hyperparameter settings.

We investigate how the ability to find valid solutions to latent space planning problems 
is affected by the hyperparameters used during the latent model learning.
We ran \latplan on the MNIST-8 domain, with
60 different hyperparameter value combinations
(all combinations of $(N, \epsilon, \beta_1, \beta_3)$ for $ N \in \{50, 100, 300\},\ p \in \{0.1, 0.5\},\ \beta_1 \in \{1, 10\},\  \beta_3 \in \{1, 10, 100, 1000, 10000\}$)
to learn 60 different latent domains, all using the AMA4+ action model \cite{AsaiKFM22}.
We then solved the 20 instances for each of these latent domains with Fast Downward using \astar with a blind heuristic.

The search results (for each domain, the number of instances for which solutions were found/valid/\coptimal) are shown in  Table \ref{tab:hyperparameters}.
With the 3 best hyperparameters $(N=300, \epsilon = 0.1, \beta_1 = 10, \beta_3 = 1)$,  $(N=300, \epsilon = 0.1, \beta_1 = 10, \beta_3 = 10)$  and $(N=300, \epsilon = 0.5, \beta_1 = 1, \beta_3 = 1)$, optimal solutions are found for all 20 instances. However, for the other 57 hyperparameter combinations, there are failures to find any plans (regardless of validity), failures to find valid plans, and/or failures to find optimal plans.
Thus, hyperparameter settings for model learning have a significant impact on the ability of \latplan to find valid plans.

Note that although previous work by showed that \latplan can find valid plans on all of the domains tested, the results  shown in this previous work (e.g., \cite[Table 11.1]{AsaiKFM22}) are for the hyperparameter configurations that yielded the highest number of valid plans {\it for each domain}, i.e., the hyperparameters were tuned for each domain \cite[p.1642]{AsaiKFM22}.

In general, we do not know {\it a priori} the hyperparameter values necessary to facilitate finding valid plans. Furthermore, hyperparameter tuning for this purpose is not possible in general, as that would require the ability to evaluate whether the plans found by the planner are valid. While we use a domain-specific ground-truth based validator in our experiments, a ground-truth domain model is not accessible in many situations where learning systems such as \latplan are necessary.
Therefore, developing a method to compensate for suboptimal model learning hyperparameter settings by improving the search component offers a significant opportunity to improve the robustness of \latplan.

\section{Plausibility-Based Heuristics (PBH)}
\label{sec:plausibility-metrics}

We propose an 
approach to estimating the plausibility of a state based
on invariance checking.
We assume that valid latent states
satisfy some invariant property
of the domain.
Consider the image-based sliding tile-puzzle or blocks world domains. If we compare, for example, start state image with the image of a valid state, although corresponding pixels differ because parts of the scene are moved by actions, we would expect that some overall statistical properties of the pixels in the images are invariant -- the locations of the pixels are changing (tile $t$ moves from location $i$ to $j$), but tiles are not being deformed/created/destroyed by the actions.

Among the domains to which \latplan has been applied so far, all of the sliding-tiles variants, 
blocks world and Towers of Hanoi satisfies this assumption.
In contrast, the LightsOut domain does not, as each action significantly changes ratio of white/black pixels.
The color version of Sokoban used in \cite{AsaiKFM22} slightly violates this assumption, as empty goal location cells have color $c_g$, differently from the other unoccupied floor cells (color $c_f$), so when the agent pushes a box (color $c_b$) from location $b$  to a goal cell $g$,  the distribution of colors before the move ($\mathit{color}(b)=c_b$, $\mathit{color}(g)=c_g$) and after the move ($\mathit{color}(b)=c_f$, $\mathit{color}(g)=c_b$) differ.

Our invariant-based approach to computing the plausibility of a latent space state $s^l$ is as follows:
Let  $r^l$ be a {\it reference state}, a latent space state which is
known/assumed to correspond to a valid state in the canonical domain.

We apply a similarity metric to ($s^l$, $r^l$) which seeks to capture
an invariance property, such that
a latent state which is more similar to $r^l$ is more likely to
correspond to a valid canonical state than a state which is less similar to $r^l$.

Regarding the choice of the reference state $r^l$, any image which is believed to be valid image (e.g., any image in $\Xtr \cup o_I \cup o_G$) is a candidate to to be used as $r^l$, but the effectiveness of the similarity metric may depend on the specific choice of $r^l$
In the experiments below, we use the reference state $r^{l} = \mathit{Encode}(o_G)$, which is the propositional latent space state corresponding to the goal image $o_G$. Investigating other choices for $r^{l}$ is future work.

Since latent propositions $P^L$ do not correspond straightforwardly to propositions $P$ of the canonical problem $T$,
it is not obvious how to apply the idea of invariant-based plausibility
to latent state vectors. Preliminary attempts to designing similarity metrics for latent state vectors have not yet been successful.
Thus, instead of directly using the latent proposition vector $s^l$
to determine the plausibility of $s^l$,
we can instead use the image corresponding to $s^l$.

To obtain the image corresponding to latent state $s^l$,
we use the $\mathit{Decode}$ function learned by \latplan
to convert $s^l$ to an image $M_{s^l} = \mathit{Decode(s^l)}$, 
where $M_{s^l}$ is an array of pixels (integers in $[0,A]$).
We then apply simple image analysis techniques to $M_{s^l}$ in order to determine the plausibility of $s^l$.
Let $H_{M_{s^l}}$ denote the {\it histogram} of an image $M_{s^l}$.
For simplicity, we write $H_{s^l}$ to mean $H_{M_{s^l}}$.
$H_{s^l}$ is an array of bins, where bin $H[b]$ is the frequency of
pixels in %
$ [ \lfloor \mathit{A}/B \rfloor \times b,   \lfloor \mathit{A}/B \rfloor \times (b+1) ]$
where $B$ is the \# of bins.

We use the standard \chisquare difference and $KL$-divergence to compare $s^l$ and the reference state $r$:

\begin{equation}
p_{\mathit{\chi^2}}(s^l) = - \sum _b \frac{\left(H_r[b]-H_{s^l}[b]  \right)^2}{H_r[b]} \tag{$\chi$-squared difference}
\end{equation}
\begin{equation}
p_{\mathit{KL}}(s^l) = - \sum _b H_r[b] \log \left(\frac{H_r[b]}{H_{s^l}[b]}\right) \tag{$\mathit{KL}$-divergence}
\end{equation}

The plausibility metrics proposed above can be used to guide a best-first algorithm such as \astar \cite{DBLP:journals/tssc/HartNR68} or Greedy Best-First Search (GBFS).
\astar uses a node evaluation function $f(n) = g(n) + h(n)$, where $g(n)$ is the cost to reach node $n$ from the initial node and $h(n)$ is the estimated cost from $n$ to a goal. GBFS uses the evaluation function $f(n) = h(n)$.
The heuristic functions \hchisquare and \hkldivergence directly use the corresponding plausibility metrics. We scale \hchisquare and \hkldivergence to an integer (as we implement these in FastDownward, which uses integer-valued heuristics).
\begin{equation}
  h_{\mathit{\chi^2}}(s^l) =  \lfloor - p_{\mathit{\chi^2}}(s^l) \rfloor
\end{equation}
\begin{equation}
  h_{\mathit{KL}}(s^l) =  \lfloor - p_{\mathit{KL}}(s^l) \rfloor
\end{equation}

Our current implementation of PBH extends the state evaluation procedure of the Fast Downward planner so that when evaluating $s^l$, it
sends a request to an external server process which uses \latplan to convert $s^l$ to an image $M_{s^l}$ and then uses OpenCV to compute an image-based plausibility metric based on 
$M_{s^l}$. 
This evaluation accounts for the vast majority of the runtime consumed by the search algorithm.
Due to the slow state evaluation rate, solving a latent PDDL instance using the current unoptimized implementation of PBH is much slower than
searching the same PDDL instance using unmodified Fast Downward.
Since the focus of this paper is on improving the ability to find valid plans,
optimization of the evaluation procedure, as well as search algorithms which do not evaluate plausibility for every state are directions for future work.

\section{Experiments}
\label{sec:experiments}

We evaluate the search performance of PBH.
We implemented PBH by extending the Fast Downward search code.
We compared:
(1) Baseline latent space search (\astar, GBFS).
using standard heuristics (the 
Blind ($h=1$) heuristic,
the admissible LMCut \cite{DBLP:conf/aips/HelmertD09} and Merge-and-Shrink \cite{DBLP:journals/jacm/HelmertHHN14} heuristics,
as well as the non-admissible FF heuristic \cite{DBLP:journals/jair/HoffmannN01}) 
and  %
(2) \astar and GBFS using plausibility-based heuristics \hchisquare and \hkldivergence (bin size $B=10$).

\subsubsection{Comparison of ability to find valid plans}
We compare the search algorithms on the following domains
(all from \citet{AsaiKFM22}, all using the AMA4+ action model):
(1) {\bf MNIST-8}: 3x3 sliding tile puzzle where the tiles have handwritten digits 0-8 from the MNIST dataset,
(2) {\bf Hanoi(4,4)}:  Towers of Hanoi (4 towers, 4 disks), 
(3) {\bf Mandrill-15}: 4x4 sliding tile puzzle where the goal is to unscramble the well-known Mandrill face image.
The learning hyperparameters were $(N=50, \epsilon = 0.1, \beta_1 = 1, \beta_3 = 1)$ for MNIST-8, 
 $(N=50, \epsilon = 0.1, \beta_1 = 1, \beta_3 = 100)$ for Hanoi(4,4), 
$(N=100, \epsilon = 0.1, \beta_1 = 1, \beta_3 = 1)$ for Mandrill-15.

\begin{table}[tb]
  \centering
  {\scriptsize
  
 \begin{tabular}{|l|r|rrrr|rrrr|}
    \hline
    \multicolumn{2}{|c}{} & \multicolumn{4}{|c|}{Hanoi(4,4)}
    & \multicolumn{4}{c|}{MNIST-8} \\
    \hline
    
    \rotatebox{90}{\parbox{0pt}{Search Alg.}}  & \rotatebox{90}{Heuristic} %
    &  \rotatebox{90}{found} & \rotatebox{90}{valid} & \rotatebox{90}{\coptimal}
    & \rotatebox{90}{length}
    &  \rotatebox{90}{found} & \rotatebox{90}{valid} & \rotatebox{90}{\coptimal}
    & \rotatebox{90}{length} \\
    \hline
    \multicolumn{10}{|c|}{Baselines}\\
    \hline
    \astar   & blind &  20 & 1 & 1 & 6.45  & 20 & 9 & 9 & 7 \\
    \astar   & lmc   &  20 & 1 & 1  & 6.45 & 20 & 6 & 6 & 7 \\
    \astar   & \mands &  20 & 2 & 2 & 6.45 & 20 & 7 & 7 & 7 \\
    \astar   & FF &  20 & 0 & 0 & 7.85     & 20 & 5 & 5 & 8.1\\
    GBFS   & blind &  20 & 1 & 1 & 6.45    & 20 & 9 & 9 & 7 \\
    GBFS   & lmc &  20 & 0 & 0 & 10.9      & 20 & 0 & 0 & 45 \\
    GBFS   & \mands &  20 & 0 & 0 & 6.8    & 20 & 4 & 4 & 8.4 \\
    GBFS   & FF &  20 & 0 & 0 & 28.9      & 20 & 0 & 0 & 57.15 \\

    \hline
    \multicolumn{10}{|c|}{PBH: Plausibility-Based Heuristics}\\
    \hline
    \astar   & \hchisquare &  20 & 15 & {\bf 8} & 7.35   & 20 & {\bf 20} & {\bf 20} & 7 \\
    GBFS   & \hchisquare &  20 & {\bf 20} & {\bf 8} & 7.7     & 20 & {\bf 20} & 6 & 21.5 \\
    \astar   & \hkldivergence &  20 & 19 & {\bf 8} & 7.65 & 20 & {\bf 20} & {\bf 20} & 7 \\
    GBFS   & \hkldivergence &  20 & {\bf 20} & 4 & 8.75  & 20 & {\bf 20} & 9 &  30.2\\
    \hline
 \end{tabular}
     } %
    \caption{
      Search Results:  Number of found/valid/\coptimal instances (out of 20) and average plan length  (optimal length = 7 for both domains).
  }
  \label{tab:search-results}
\end{table}

First, we compared the search performance 
on the MNIST-8 and Hanoi(4,4) domains (20 instances each)
Table \ref{tab:search-results} shows the results (``found'': latent space plan found, ``valid'': plan is valid, ``\coptimal'': plan is \coptimal).
Although baseline \astar and GBFS find latent space plans, many of them are not valid.
Note that the average plan lengths include invalid plans which can be either longer or shorter than valid plans, so the average lengths differ from the optimal valid length (7) even for \astar using admissible heuristics (blind, lmc, M\&S).

PBH (both \hchisquare and \hkldivergence) found significantly more valid plans than the baselines.
In particular, PBH successfully, reliably finds valid solutions for Hanoi(4,4), on which previous work had failed to find any valid solutions \cite{AsaiKFM22}.

On Mandrill-15, we ran a subset of the algorithm configurations, as it requires significantly more runtime than Hanoi(4,4) and MNIST-8.
As explained in Section \ref{sec:plausibility-metrics} the runtime of our current unoptimized implementation of PBH is very slow (state evaluation rate is less than 100 nodes per second for Mandrill-15, and the average runtime is 8.3 hours).
The results were as follows:
(a) \astar with blind heuristic: 20 found, 15 valid, 15 \coptimal, 
(b) \astar with landmark cut: 20 found, 14 valid, 14 \coptimal, 
(c) \astar with merge-and-shrink: 20 found, 15 valid, 15 \coptimal, 
(d) GBFS with blind heuristic:  20 found, 15 valid, 15 \coptimal, 
(e) \astar with \hkldivergence: 20 found, 20 valid, 20 \coptimal.
Thus, PBH significantly improves the ability to find valid solutions on Mandrill-15.

\subsubsection{Search Behavior}

\begin{figure}[tb]
  \includegraphics[height=1.5in,width=0.9\linewidth]{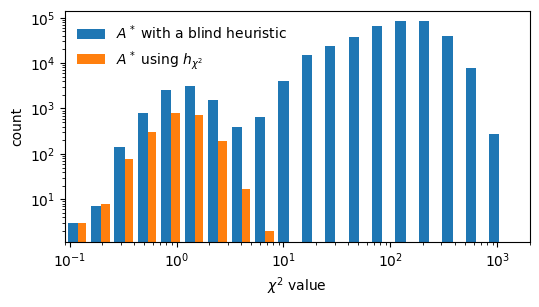}
  \includegraphics[height=1.5in,width=0.9\linewidth]{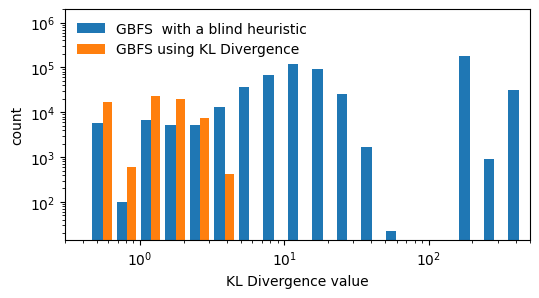}
  \caption{Top: $h_{\chi^2}$ values of states expanded during search on MNIST-8 puzzle (20 instances).
    Bottom: $h_{\textit{KL}}$ values of states expanded during search on Hanoi(4,4) (20 instances).}
 \label{fig:p-value-histogram}
\end{figure}

In order to compare the search behavior of PGH vs. baseline search,
we compare the plausibility scores of all states expanded the search on all 20 instances of MNIST-8 and Hanoi(4,4) by a baseline search (\astar/blind  for MNIST, GBFS/blind for Hanoi vs. PBH (\astar/\hchisquare for MNIST, GBFS /\hkldivergence for Hanoi).
For both baseline and PBH, we stored the plausibility values of the expanded states (the baseline computes but ignores them during search).
Figure \ref{fig:p-value-histogram} shows that the plausibility
values of the states expanded by the baseline are much more widely distributed than those of the states expanded by \hchisquare and \hkldivergence, confirming that search with PBH focuses on highly plausible states.

\section{Conclusion}
\label{sec:conclusion}

In a system such as \latplan which is intended to operate autonomously
in unknown environments without a teacher who can identify whether a plan is valid or invalid, flawed learned models which lead to invalid plans poses a significant challenge to the utility and robustness of the system.
This work showed that even if a learned, domain model is flawed,
the search algorithm can compensate %
by avoiding suspicious (low plausibility) states,  resulting in significantly improved ability to find valid plans.
Although PBH was evaluated on \latplan, we believe
the overall approach may %
be useful in similar planning systems where learned problem representations have spurious states.

Our approach is domain-independent.
The plausibility metrics do not use any domain-specific notions, e.g., there are no ``tile/disc detectors''. Instead, we rely on properties common to some subset of \latplan domains, i.e., invariance properties of the histogram of the image representation of valid states, and our plausibility metrics
successfully improve search results in multiple domains. %
Developing plausibility metrics for other types of domains, e.g., LightsOut, is a direction for future work.

This paper evaluated using plausibility metrics as heuristic evaluation functions to guide search instead of standard distance-based heuristics.
In some cases, this  may result in degraded search performance or solution quality compared to distance-based heuristics. However, there are many situations where an invalid plan may be no better than finding a poor quality solution or failing to find a plan altogether, so in this paper we focus on prioritizing validity.
Future work will investigate combining plausibility with distance based heuristics by using plausibility as a tie-breaking criterion \cite{DBLP:conf/aips/AsaiF17} or by considering both plausibility and distance in the heuristic evaluation function in order to improve solution quality and search performance.

This work opens avenues for future work on
search strategies as complementary approaches to improvements to the learning
component in domain-learning systems.
This paper explored the use of plausibility metrics to evaluate the plausibility of a single state during search.
Extending the notion of plausibility to paths, e.g., prioritizing paths according to plausibility, may further improve search behavior. %
In this paper, we focused on improving the ability to find valid plans by straightforwardly applying plausibility as a heuristic evaluation function 
in standard \astar and GBFS search.
However, PBH incurs significant runtime costs while computing the plausibility of a state.
Development of more sophisticated search strategies
 which seek to reduce the number of plausibility computations while still avoiding invalid plans is a direction for future work.

\section*{Acknowledgments}
Thanks to Masataro Asai for helpful discussions and assistance with the \latplan source code and sample model files. Thanks to the anonymous reviewers for helpful comments.
\fontsize{9.0pt}{10.0pt}
\bibliography{reference}

\end{document}